# Do Large Language Models know who did what to whom?

Joseph M. Denning[1]* (josephdenning@ucla.edu)
Xiaohan (Hannah) Guo[2]* (hannahguo@uchicago.edu)
Bryor Snefjella[1]
Idan A. Blank[1] (iblank@psych.ucla.edu)
[1]Department of Psychology, 502 Portola Plaza, Los Angeles, CA 90095 USA
[2]Department of Psychology, 5848 S University Ave, Chicago, IL 60637 USA
*Equal contribution

## Abstract (149 words)

Large Language Models (LLMs) are commonly criticized for not "understanding" language. However, many critiques focus on cognitive abilities that, in humans, are distinct from language processing. Here, we instead study a kind of understanding tightly linked to language: inferring "who did what to whom" (thematic roles) in a sentence. Does the central training objective of LLMs—word prediction—result in sentence representations that capture thematic roles? In two experiments, we characterized sentence representations in four LLMs. In contrast to human similarity judgments, in LLMs the overall representational similarity of sentence pairs reflected syntactic similarity but not whether their agent and patient assignments were identical vs. reversed. Furthermore, we found little evidence that thematic role information was available in any subset of hidden units. However, some attention heads robustly captured thematic roles, independently of syntax. Therefore, LLMs can extract thematic roles but, relative to humans, this information influences their representations more weakly.

**Keywords:** large language models; comprehension; thematic roles; representational similarity

# Introduction

Large Language Models (LLMs) have achieved unprecedented success at natural language processing. Their success demonstrates the power of statistical learning over strings of linguistic forms (Contreras Kallens et al., 2023; Piantadosi, 2023): by merely learning to predict the next word (or a missing word) in a text, LLMs develop the ability to produce texts that conform to the syntactic rules of a language (e.g., McCoy et al., 2023). Indeed, the internal representations and next-word predictions of LLMs suggest that these systems have acquired many complex grammatical generalizations (for reviews, see Chang and Bergen, 2024; Futrell and Mahowald, 2025; Linzen and Baroni, 2021). As a result, LLMs have been proposed as models of human language processing (Blank, 2023) and are being used to predict both behavioral and neuroimaging data (Antonello et al., 2021; Caucheteux and King, 2022; Evanson et al., 2023; Hosseini et al., 2022; Merkx and Frank, 2020; Oh and Schuler, 2022; Schrimpf et al., 2021; Tang et al., 2024; Tuckute et al., 2024; Wilcox et al., 2020). Nonetheless, humans do not use grammar as an end in itself, but rather as an intermediate step in mapping linguistic input forms (e.g., phonological or orthographic information) onto meaning (semantics; Jackendoff and Wittenberg, 2017). Therefore, a key question is whether LLMs can extract meaning from their textual input. More specifically: what kinds of meaning can be acquired merely from learning to predict the next (or a missing) word?

The general question of semantics in LLMs is hotly debated. Whereas the behavior and internal activity of LLMs exhibit some signatures of semantic representations (for a review, see (Chang and Bergen, 2024; Pavlick, 2022), these models are often criticized for not truly "understanding" language (e.g., Bender & Koller, 2020). The critiques use various definitions of "understanding". Some critics claim that LLMs do not have "grounded" knowledge that links linguistic meaning to non-linguistic experience (Bender and Koller, 2020; Bisk et al., 2020; Merrill et al., 2021); others claim that LLMs lack common sense, i.e., intuitive theories about how the world works (Sinha et al., 2019; Ullman, 2023); yet others claim that LLMs do not reason logically (Ettinger, 2020; Wu et al., 2023). All such criticisms conclude that LLMs do not understand language like humans do.

However, these critiques are valid only to the extent that they rely on accurate notions of language comprehension in humans. Humans are, of course, able to relate linguistic input to non-linguistic experiences, evaluate such input against their common sense and prior knowledge, and use it for logical inferences. But these capacities are kinds of "thinking", not language (Mahowald, Ivanova, et al., 2024): in the human mind, the cognitive systems that support sensorimotor or affective processes, common sense reasoning, and logical inferences are functionally distinct from the system that analyzes linguistic input (Fedorenko et al., 2024; Fedorenko and Varley, 2016). Whereas linguistic processing is a prerequisite for, e.g., evaluating whether a sentence is consistent with common sense, these two processes are dissociable, and a failure to carry out the latter does not demonstrate a failure to carry out the former (by analogy, if a human fails to paint a copy of a picture, that does not mean they failed to visually perceive that picture). Given that the mind appears to dedicate a system to linguistic processing per se, a fair yet critical bar for LLMs to pass is semantic processing that is language-internal or, at least, closely tied to language (e.g., Piantadosi and Hill, 2022, but see Jackendoff, 2012).

Here, we test such a case of understanding: using the structure of a sentence that describes a simple event to figure out "who did what to whom", a process called "thematic role assignment" (Fillmore, 1968; Jackendoff, 1972; Levin & Hovav, 2005; see also Dryer, 2002; Fenk-Oczlon & Fenk, 2008; Greenberg, 1963; Kiparsky, 1997; Koplenig et al., 2017; Levshina, 2020, 2021; Sinnemäki, 2008; for a review, see Rissman & Majid, 2019). Specifically, we focus on the mapping from the grammatical roles of a verb's "subject" and "object" to the thematic roles of an action's "agent" and "patient". This mapping is variable across syntactic structures: in an active sentence like "the tiger punched the panther" (**Figure 1**), the grammatical subject (tiger) is the agent, and the grammatical object (panther) is the patient; however, in a passive sentence, like "the tiger was punched by the panther", the roles are reversed: the grammatical subject is still the tiger, but it is now the patient being punched, whereas the grammatical object—still panther, as in the previous active sentence—is now the agent doing the punching. As this example demonstrates, two sentences with different grammatical constructions (active vs. passive) can have the same thematic role assignments. Alternatively, two sentences with the same grammatical construction can have opposite thematic role assignments, e.g., "the tiger punched the panther" vs. "the panther punched the tiger" (**Figure 1**). Inferring who did what to whom requires combining several types of syntactic information (linear order, construction), at least in the absence of prior semantic information about which noun is a more plausible agent (Caramazza and Zurif, 1976; Mahowald et al., 2023).

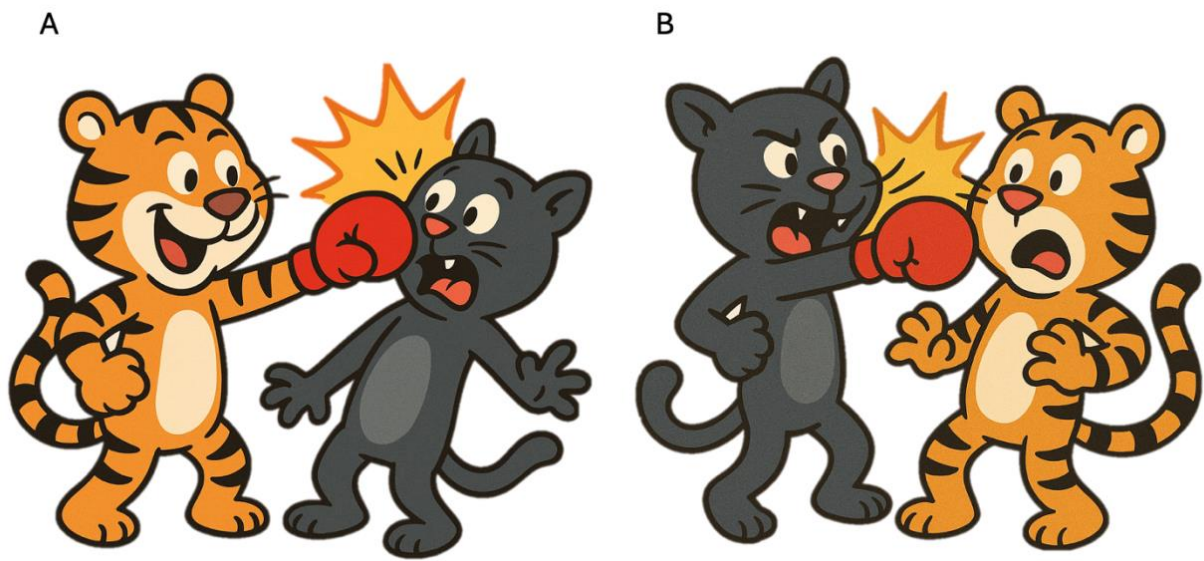

**Figure 1:** Determining thematic roles is important for distinguishing similar events. (A) The event depicted on the left can be described by "the tiger punched the panther" or "the panther was punched by the tiger". (B) The event depicted on the right can be described by "the panther punched the tiger" or "the tiger was punched by the panther". Images generated from DALL-E 3, April 2025.

Thematic role assignment is often invoked as an important component of language processing in psycholinguistic theories (for a review, see: Rissman & Majid, 2019), unlike many other tasks used to test LLMs, such as logical entailment or common-sense reasoning (e.g., Srivastava et al., 2022; Wang et al., 2018). It is central to the syntax-semantics interface (Goldberg, 2019, 1995) and becomes available rapidly to guide incremental comprehension (Altmann, 1999; Gennari and MacDonald, 2008). However, it appears to be dissociable from syntactic processing per se (Caramazza and Miceli, 1991; Chatterjee et al., 1995; Ziegler and Snedeker, 2018), and thus the evidence for syntactic representations in LLMs does not trivially predict that thematic role assignment would be successful.

In the human brain, thematic role assignment engages the Core Language Network (Ivanova et al., 2021), a system selective for high-level language processing (Fedorenko et al., 2024, 2011), including the extraction of meaning from sentences (Fedorenko et al., 2016; Regev et al., 2024). Whereas thematic role assignment also recruits a-modal regions located outside the Core Language Network, this recruitment critically relies on task demands beyond role assignment itself: unlike the Core Language Network, these other regions are overall not sensitive to linguistic meaning (Ivanova, 2022; see also Frankland & Greene, 2020; Wang et al., 2016). Thus, understanding "who did what to whom" in a sentence appears to predominately (or, at least, heavily) rely on computations that are linguistic in nature, and thus provides an appropriate test for LLMs.

In two experiments, we test whether training LLMs on a word-prediction objective results in representations that reflect thematic roles. To this end, we use LLMs that are pre-trained on this objective, without any further fine-tuning on other objectives. Five matters about this rationale are worth emphasizing. First, LLMs can be directly trained on the specific task of thematic role assignment (also called "semantic role labeling") via supervised learning, wherein training sentences are paired with correct role labels for different words. However, our question is whether the broader training objective of word prediction suffices for this purpose. Word prediction is the consensus objective for (pre)training LLMs and, as mentioned above, such LLMs are treated as "general language processors" that are, e.g., compared to human behavior and brain activity. Our study asks whether these models "understand" sentences in the minimal sense of thematic role assignment.

Second, many state-of-the-art LLMs, like ChatGPT (OpenAI, 2022) or GPT4 (OpenAI, 2023), are not only (pre)trained on word prediction but are also fine-tuned using "reinforcement learning from human feedback" (RLHF; Christiano et al., 2017; Ouyang et al., 2022). In RLHF, an LLM receives input about human preferences and learns a policy for generating text that is maximally aligned with those preferences. Such "evaluative" input—information about how humans would like the LLM to respond to their queries—constitutes non-linguistic information according to our construal of the distinction between language and thought. LLMs trained with RLHF are therefore outside the scope of the current work, give that we are asking whether meaning can be learned just from information about the distribution of words and sentence structures.

Third, prior work has demonstrated that the internal representations of LLMs contain information about thematic roles (e.g., Tenney, Das, et al., 2019; Tenney, Xia, et al., 2019). However, the presence of such information is often tested using corpora derived from natural text (e.g., Carreras & Màrquez, 2005; Pradhan et al., 2013), which likely contains few challenging examples: in many sentences, thematic roles might be assigned based on heuristics or shortcuts (Mahowald et al., 2023). Our study instead relies on stimuli that are specifically designed to be less susceptible to such heuristics: both experiments use "reversible" sentences in which both agent and patient are equally likely to produce the action, and Experiment 2 uses a wide range of grammatical constructions. We chose this approach because other linguistic capacities that appear robust when tested on corpora "from the wild" can break down when tested on carefully crafted stimuli, like those used in controlled psycholinguistic experiments conducted with humans (e.g., Chaves and Richter, 2021; Glockner et al., 2018; McCoy et al., 2019; Rosenman et al., 2020; Sinha et al., 2021).

Fourth, testing whether LLMs map syntax onto thematic roles assumes that a representation of syntactic structure is available to LLMs. However, syntactic processing in LLMs still falls short of humans (Aina and Linzen, 2021; Marvin and Linzen, 2018; Van Schijndel and Linzen, 2021); and even when LLMs do capture the structure of sentences, they might rely on "tricks" that differ from the rich, systematic linguistic principles guiding humans (Chaves & Richter, 2021; McCoy et al., 2019; Sinha et al., 2021). Still, such characterizations of LLMs are often based on sentences with quite complex structures. There is wide implicit agreement that LLMs do capture the structure of simple sentences like "the tiger punched the panther", which are the stimuli used in Experiment 1. Additionally, LLMs appear to be able to represent syntactic information "when it matters" (Papadimitriou et al., 2022).

Fifth, and finally, our approach involves investigating the internal representations of LLMs rather than prompting them (e.g., directly asking the models "what is the agent of this sentence?" or "do these sentences have the same thematic roles?"). We use this probing approach because a prompting task would require that these models not only assign thematic roles, but also that they understand the question being asked of them and understand what words like "agent" or "thematic roles" mean (meta-linguistic abilities). Probing the models directly allows us to get around these task demands, and previous work shows that these distinct approaches can lead to differing results (Hu et al., 2024; Hu and Frank, 2024; Hu and Levy, 2023). Even if a LLM can answer an explicit prompt correctly, such accuracy does not demonstrate that the model has a robust and generalizable internal encoding of thematic roles. Instead, the LLM might "simulate" understanding via shallow heuristics, memorization of templates, or other distributional patterns in its training corpora. Moreover, prompts measure performance in a specific task context. Studying LLM knowledge across sufficiently many contexts and prompt variations is practically challenging and, moreover, performance that is not stable across such variations would be difficult to interpret (are some prompts "bad" or is the LLM brittle?). Rather than quantifying performance via LLM outputs, internal representations quantify something closer to the underlying "competence": abstract latent structure that is apparent even when not explicitly prompted.

Our approach is therefore appropriate for testing whether LLMs implicitly assign roles like "agent" and "patient" to event participants in a sentence. Our main analyses rely on the following paradigm: we feed different sentences to these LLMs, extract the resulting sentence representations—in the form of patterns of activity in hidden layers—and quantitively characterize to what extent they are influenced by thematic role information. Specifically, we generate sentences that either (i) share thematic role assignments but differ in syntax (e.g., "the tiger punched the panther" and "the panther was punched by the tiger"), or (ii) have opposite thematic role assignments but share syntax (e.g., "the tiger punched the panther" and "the panther punched the tiger"). We test whether sentence pairs of type (i) are more similar to one another than pairs of type (ii), which would be expected if word prediction suffices for LLMs to learn something akin to thematic roles, and if this aspect of meaning plays a larger role than syntax in internal representations. This is a simple form of representational similarity analyses in fMRI (Kriegeskorte et al., 2008; it is perhaps most similar to the fMRI method "multi-voxel pattern analysis", Norman et al., 2006, see other work that uses a similar method: Cong, 2024; Nicolas and Caliskan, 2024). We also test to what extent human similarity ratings of the same sentence pairs reflect thematic role assignments. If LLMs are good models of human language processing, then the influence of thematic role information on their representations should have similar strength to its influence on human judgments.

## Data Availability

Code and data for all experiments can be found at https://osf.io/t4raz/.

## Experiment 1

### Methods

**Stimuli.** Stimuli were based on a fMRI experiment by Fedorenko et al., (2020) and generated in two steps: First, we created 94 "base" active, transitive sentences describing a two participant event, such as "the lawyer saved the author". Then, we edited each base sentence to create four versions that changed its thematic role assignments (and hence the semantics) and/or its syntax, as outlined in **Table 1**: (A) same semantics, same syntax ($SEM_s$-$SYNT_s$): a "control" version where nouns and verbs are replaced by near-synonyms (extracted using the static word embedding GloVe; Pennington et al., 2014), maintaining the active structure and the thematic role assignments ("the attorney rescued the writer"); (B) same semantics, different syntax ($SEM_s$-$SYNT_d$): a critical condition where the sentence is converted to passive voice while maintaining its base words and thematic role assignments ("the author was saved by the lawyer"); (C) different semantics, same syntax ($SEM_d$-$SYNT_s$): the other critical condition where the agent and the patient are swapped, thus changing thematic role assignments while maintaining its base words and active structure ("the author saved the lawyer"); and (D) different semantics, different syntax ($SEM_d$-$SYNT_d$): another "control" version where nouns and verbs are replaced with near-synonyms, the agent and patient are swapped, and the sentence is converted to passive voice, resulting in a sentence that is maximally different from the base ("the attorney was rescued by the writer").

**Table 1.** Sample Experiment 1 materials, compared to the base sentence "The lawyer saved the author".

|  | Same Structure (active) | Different Structure (passive) |
|---|---|---|
| **Same Meaning** | (A) The attorney rescued the writer | (B) The author was saved by the lawyer |
| **Different Meaning** | (C) The author saved the lawyer | (D) The attorney was rescued by the writer |

To the extent that LLMs should capture sentence meaning, and that thematic roles are one of the basic components of meaning, LLMs should represent sentences with the same thematic role assignments (and thus similar meanings) more similarly than sentences with opposite assignments (and thus less similar meanings): the base sentence would be most similar to condition $SEM_s$-$SYNT_s$, followed by $SEM_s$-$SYNT_d$ (where thematic roles are still the same, but the syntax is different), then $SEM_d$-$SYNT_s$ , then $SEM_d$-$SYNT_d$. (Note that sentence similarities might be strongly influenced by whether two sentences share the exact same nouns and verbs or, instead, use different words even if they are near synonyms; regardless of this issue, our critical comparison is between the similarity of the base to $SEM_s$-$SYNT_d$ and its similarity to $SEM_d$-$SYNT_s$). If, however, LLMs fail to represent event meaning, similarities would only / mostly reflect whether sentences share structure, regardless of "who did what to whom". In this case, the base sentence would be more similar to $SEM_d$-$SYNT_s$ than to $SEM_s$-$SYNT_d$. These predictions are depicted in **Figure 2**.

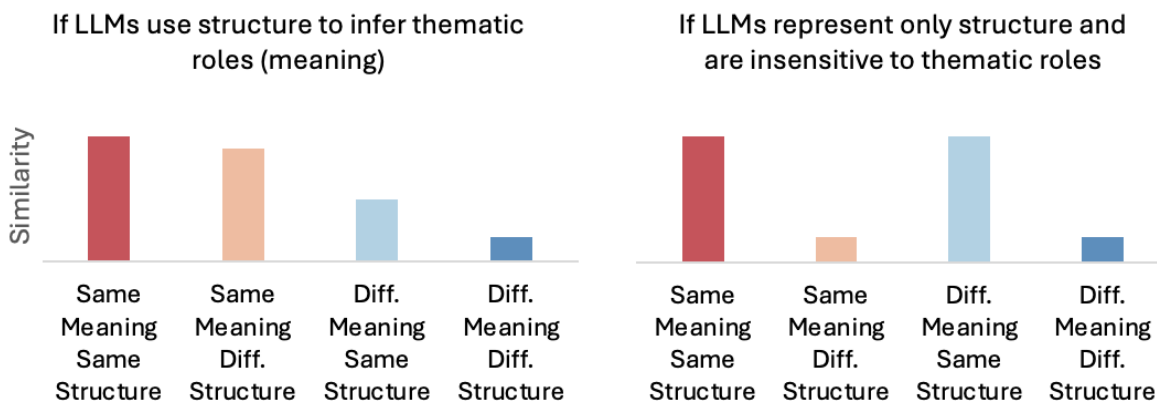

**Figure 2.** Expected similarity patterns between sentence pairs under two hypotheses: LLMs may represent the structure of sentences and use it to infer thematic roles (left), or they may represent structure while being insensitive to thematic roles (right).

**Large language models.** We studied sentence representations in BERT (110M parameters; Devlin et al., 2018), GPT2-Small (117M parameters; Radford et al., 2019), Llama 2 (7B parameter version; Touvron et al., 2023), and Persimmon (8B parameter version; (Elsen et al., 2023) as implemented in HuggingFace. BERT and GPT2-Small have 12 layers, each with 768 hidden units and 12 attention heads. Llama2 has 32 layers, each with 4096 hidden units; and Persimmon has 36 layers each with 4096 hidden units. BERT is bidirectional, and all other models are unidirectional.

**Evaluating representational similarities.** For each layer in each LLM, we extracted a representation of each sentence. A sentence representation consists of a distributed pattern of activity across hidden units. Specifically, these activities were extracted for the [CLS] token in BERT, and the '.' token in GPT2, Llama2, and Persimmon (Devlin et al., 2018; May et al., 2019; Schrimpf et al., 2021). We also investigated mean pooled activations (averaging across tokens), as well as the verb token in BERT (unlike BERT, in unidirectional LLMs the representations of tokens other than the final one cannot capture information about the entire sentence). The results are for these analyses, reported in the supplementary materials, do not qualitatively differ from our main analyses.

For each sentence set, we compared the representation of the base sentence and each other condition via the cosine similarity measure (Cassani et al., 2023). Because such similarities might be influenced by a small subset of hidden units with, e.g., very high activations across all sentences (Timkey and van Schijndel, 2021), prior to computing similarities we normalized each hidden unit's activations relative to that unit's average and standard deviation across a large set of sentences (COCA; Davies, 2009). Cosine similarities were Fisher-transformed to render their distribution closer to Gaussian and ameliorate bias in averaging them across stimuli (Silver & Dunlap, 1987). We contrasted the four conditions in terms of their similarity to the base sentence using a non-parametric, one-way repeated-measures ANOVA (Friedman test). Significant results were followed by pairwise post-hoc, two-tailed Wilcoxon signed rank tests (Bonferroni corrected).

**Human judgments.** Human judgments about our stimuli are required as a standard against which to evaluate LLMs (Arana et al., 2023). Humans can identify who did what to whom, even based on grammatical cues alone (without, e.g., plausibility information), upon careful contemplation. For example, identifying the agent and patient in "the ball was kicked by the boy" is not difficult, as common sense makes it much more plausible that the boy is the agent and the ball is the patient. But it is also possible, with some contemplation, to identify the agent and patient in sentences that do not contain such plausibility information, as "the chef was pushed by the painter". Even though either a chef or a painter can push one another, we can rely on the syntax to identify that the painter is the agent and the chef is the patient. Still, it remains unclear how automatically and accurately we infer thematic roles based on grammatical cues alone (versus plausibility cues) and without explicit instructions to closely attend to sentence structure (Ferreira & Lowder, 2016).

To this end, we collected behavioral judgments from 120 participants, recruited via UCLA's participant recruitment system ($n$=3 removed due to missing responses) (mean±SD age: 20.19y±2.72; 91 female, 25 male, 1 other). The study was approved by the university's Institutional Review Board.

In an online experiment, participants rated pairs of sentences for their similarity, using a sliding scale between 1 (completely different) and 100 (identical). To minimize the chances that participants detect the distinctions between the conditions in **Table 1** and develop an artificial strategy for solving the task, each participant made only one judgment per condition (i.e., rated the similarity between a single sentence from that condition and its corresponding base sentence). Stimuli across the four conditions came from distinct sets (no base sentence was read more than once by a participant). We created 24 experimental lists, each consisting of 4 sentence pairs and shown to 5 participants. To mask the purpose of the study, these pairs were interleaved among 5 other pairs where similarity did not require close attention to sentence structure and event roles, because it could instead be derived from general common sense (e.g., we expected two sentences about food, such as "the boy baked a cake" and "the girl devoured the sandwich", to be more similar to one another compared to two sentences on distinct topics, such as "the soldiers stormed the camp" and "the spectators enjoyed the movie").

We $z$-scored similarities within each participant. We then attempted to fit a linear, mixed-effects model predicting sentence similarity from condition, with random intercepts by participant and/or stimulus set: Similarity ~ Condition + (1 | Participant) + (1 | Set). These models did not converge and showed little variance across participants (due to $z$-scoring) and across sets. Therefore, we ran a fixed-effects model predicting similarity between sentences from condition (Similarity ~ Condition).

Both humans and LLMs may succeed in this task. Alternatively, LLMs and humans might err in similar ways, with both failing to reliably assign thematic roles. But if LLMs and humans diverge in their performance patterns, it would suggest that linguistic representations in LLMs are, in some crucial ways, different from those in human minds.

## Results

We report results for the last layer in each model, but the critical findings hold across layers (**Supplementary Figure 1**). Below, we use "similar" to mean "similar to one another".

**LLMs.** For all four LLMs, we found a significant difference between cosine similarities across conditions (**Figure 3A-D**, **Table 2**) (BERT: $\chi^2_{(3)}$=223.0, $p$<$10^{-47}$, Bonferroni-corrected here and below; GPT2: $\chi^2_{(3)}$=243.17, $p$<$10^{-51}$; Llama2: $\chi^2_{(3)}$=119.7, $p$<$10^{-25}$; Persimmon: $\chi^2_{(3)}$=136.9, $p$<$10^{-28}$). Most critically, post-hoc tests revealed a surprising finding: a pair with opposite meanings (but the same syntax) was more similar than a pair with the same meaning (but different syntax) ($SEM_d$-$SYNT_s$ > $SEM_s$-$SYNT_d$). This finding demonstrates that syntax exerts a stronger influence on LLM representations than thematic roles do.

**Table 2.** Results of Experiment 1 for each LLM.[a]

| Comparison | BERT | GPT2 | Llama2 | Persimmon |
|---|---|---|---|---|
| $SEM_d$-$SYNT_s$ > $SEM_s$-$SYNT_d$ | $z$=8.34, $p$ = $10^{-15}$ $CI_{95\%}$ = [0.95, 1] | $z$=8.23, $p$<$10^{-14}$ $CI_{95\%}$ = [0.88, 0.98] | $z$=6.86, $p$<$10^{-10}$ $CI_{95\%}$ = [0.77, 0.91] | $z$=7.14, $p$<$10^{-11}$ $CI_{95\%}$ = [0.87, 0.97] |
| $SEM_s$-$SYNT_s$ > $SEM_d$-$SYNT_d$ | $z$=5.11, $p$=$10^{-5}$ $CI_{95\%}$ = [0.66, 0.84] | $z$=4.50, $p$<$10^{-4}$ $CI_{95\%}$ = [0.61, 0.79] | $z$=4.05, $p$<$10^{-3}$ $CI_{95\%}$ = [0.60, 0.79] | $z$=5.16, $p$<$10^{-5}$ $CI_{95\%}$ = [0.65, 0.83] |
| $SEM_s$-$SYNT_s$ > $SEM_d$-$SYNT_d$ | $z$=8.34, $p$ = $10^{-15}$ $CI_{95\%}$ = [0.89, 0.99] | $z$=8.33, $p$<$10^{-16}$ $CI_{95\%}$ = [1, 1] | $z$=7.13, $p$<$10^{-11}$ $CI_{95\%}$ = [0.73, 0.89] | $z$=6.47, $p$<$10^{-9}$ $CI_{95\%}$ = [0.73, 0.89] |
| $SEM_s$-$SYNT_d$ > $SEM_s$-$SYNT_s$ | $z$=6.34, $p$ = $10^{-8}$ $CI_{95\%}$ = [0.71, 0.87] | $z$=8.19, $p$<$10^{-14}$ $CI_{95\%}$ = [0.95, 1] | $z$=4.40, $p$<$10^{-4}$ $CI_{95\%}$ = [0.57, 0.76] | $z$=3.48, $p$= .003 $CI_{95\%}$ = [0.53, 0.73] |
| $SEM_d$-$SYNT_s$ > $SEM_d$-$SYNT_d$ | $z$=8.41, $p$ = $10^{-15}$ $CI_{95\%}$ = [1, 1] | $z$=8.32, $p$<$10^{-15}$ $CI_{95\%}$ = [1, 1] | $z$=7.85, $p$<$10^{-13}$ $CI_{95\%}$ = [0.85, 0.97] | $z$=7.85, $p$<$10^{-13}$ $CI_{95\%}$ = [0.91, 0.99] |

| SEM$_d$-SYNT$_s$ > SEM$_s$-SYNT$_s$ | $z$=8.41, $p$ = $10^{-15}$ CI$_{95\%}$ = [0.97, 1] | $z$=8.33, $p$<$10^{-15}$ CI$_{95\%}$ = [1, 1] | $z$=7.05, $p$<$10^{-10}$ CI$_{95\%}$ = [0.76, 0.9] | $z$=7.00, $p$<$10^{-10}$ CI$_{95\%}$ = [0.85, 0.97] |
|---|---|---|---|---|

[a] 95% confidence intervals show the probability that, for a randomly sampled stimulus set, the difference denoted on the left-most column would be observed in the right direction. Intervals were estimated using 5000 bootstrap samples.

For all LLMs, post-hoc tests also revealed that (1) the two control conditions differed from one another as expected (SEM$_s$-SYNT$_s$ > SEM$_d$-SYNT$_d$); (2) a pair with different syntax but shared meaning was more similar than the maximally different control pair (SEM$_s$-SYNT$_d$ > SEM$_d$-SYNT$_d$) but was also more similar than the maximally similar control pair (SEM$_s$-SYNT$_d$ > SEM$_s$-SYNT$_s$). The latter effect perhaps reflects the fact that SEM$_s$-SYNT$_d$ sentences shared the same nouns and verbs with the base sentence, whereas SEM$_s$-SYNT$_s$ instead consisted of near synonyms; and (3) a pair with the same syntax but different meaning was more similar than the maximally different pair (SEM$_d$-SYNT$_s$ > SEM$_d$-SYNT$_d$), but also more similar than the maximally similar pair (SEM$_d$-SYNT$_s$ > SEM$_s$-SYNT$_s$). Again, this latter effect perhaps reflects lexical overlap between pairs of sentences.

**Human Judgments**. A fixed-effects model predicting similarity between sentences from condition had an adjusted $R^2$ of 0.25, $F_{(4,458)}$=40.01, $p$<.001 (**Figure 3E**). Most critically, post-hoc tests revealed that, a pair with the same meaning (but different syntax) was more similar than a pair with opposite meanings (but the same syntax) (SEM$_s$-SYNT$_d$ > SEM$_d$-SYNT$_s$, $z$=9.27, $p$<$10^{-15}$). This pattern is the opposite of what was found for LLMs, and demonstrates that thematic roles exert a stronger influence on human similarity judgments than syntax does.

Post-hoc tests also found that: (1) the two control conditions differed from one another in the expected direction ($z$=7.83, $p$<$10^{-12}$); (2) a pair with the same meaning but different syntax was more similar than the maximally different control pair (SEM$_s$-SYNT$_d$ > SEM$_d$-SYNT$_d$, $z$=8.13, $p$<$10^{-15}$), but did not differ from the maximally similar control pair (SEM$_s$-SYNT$_d$ vs. SEM$_s$-SYNT$_s$, $z$=0.29, $p$=1); and (3) unlike in LLMs, but consistent with a strong influence of thematic roles, a pair with different meanings but shared syntax was less similar than the maximally similar pair (SEM$_d$-SYNT$_s$ < SEM$_s$-SYNT$_s$, $z$=8.96, $p$<$10^{-15}$); and, despite sharing syntax, this pair did not differ from the maximally different pair (SEM$_d$-SYNT$_s$ vs. SEM$_d$-SYNT$_d$, $z$=1.15, $p$=1).

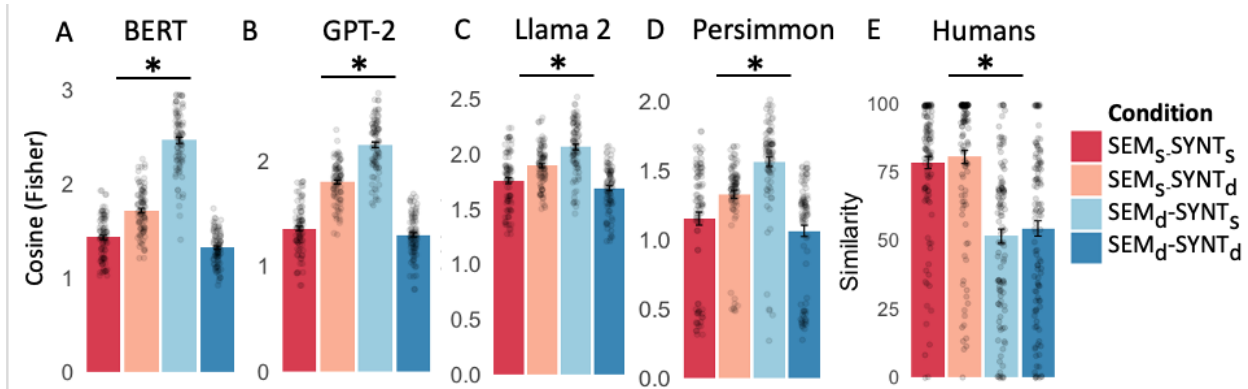

**Figure 3**: Results of Experiment 1. Similarity of the base sentences to sentences in each condition for (A) BERT, (B) GPT2, (C) Llama2, (D) Persimmon, and (E) humans. For LLMs, data are shown for the last hidden layer, and cosine similarities are Fisher-transformed. Error bars show standard error of the mean; dots show individual stimulus sets (for panels A-D) or trials (for panel E).

We did not directly compare humans and LLMs, because the similarity judgments in these two datasets are on different scales (Likert vs. cosine). However, we analyzed LLM data for the subset of 24 stimulus sets for which we collected behavioral data, and the comparison between the two critical conditions was still significant ($p<.001$ for all models). Even without a direct comparison, we emphasize that human similarity judgments are governed by thematic role assignments and go in the opposite direction from LLM representational similarities, which are governed by syntax (as evidenced by the comparison of $SEM_s$-$SYNT_d$ and $SEM_d$-$SYNT_s$).

## Experiment 2

A cognitively plausible representation of sentences should feature thematic roles as a main component (Rissman and Majid, 2019; Ziegler and Snedeker, 2018); for this reason, Experiment 1 quantified LLM representations as distributed activity patterns across all hidden units in a given layer. However, thematic roles might instead be encoded by a small subset of units (with other units representing unrelated information, e.g., lexical semantics). Thus, even if LLMs do not emphasize thematic roles—as evidenced in Experiment 1—perhaps they still extract this information, i.e., they have the capacity for representing this information, like humans do.

Experiment 2 thus addressed two questions. First, we asked: are thematic roles represented anywhere among LLM hidden units, even in a small subset of them? If this were the case, the analyses of Experiment 1 may not have been able to discover this more "localized" representation. To evaluate this possibility, we again used pairs of sentences that either shared the same agent and the same patient or had reversed thematic role assignments. We first extracted activity patterns across all hidden units for each sentence, and then submitted paired representations to an algorithm that tried to find any information, in any subset of units, that could classify whether that pair of sentences shared common thematic role assignments or not.

Our second question was whether thematic role information was available in components of LLMs other than hidden units, namely, in the attention heads. Attention heads are components that influence how the representation of each word (or, more generally, token) is transformed from one layer of hidden units to the next. Specifically, each head determines how words in a sentence are related to one another; these relations reflect various sources of information, which are inferred automatically during training, and might include semantics, syntax, linear position, etc. To this end, an attention head assigns numerical weights between words (i.e., specifying how much a word should "attend" to another): in bidirectional transformers (like BERT), each word attends to every other word and to itself, whereas in unidirectional transformers (like GPT2, Llama2, and Persimmon), each word attends only to itself and to previous words (i.e., it cannot "look ahead"). The representation of a given word then becomes a weighted linear combination of all the words it attends to, based on their respective attention weights. Different heads assign different weights, producing distinct word representations (i.e., linear combinations), which are combined across all heads in a given layer and further transformed before being passed on to the next set of hidden units. To test whether any attention heads capture thematic role information, for each head we extracted attention weights between content words in a sentence, and used the same classification procedure described above on pairs of sentences that had either shared or reversed thematic role assignments.

## Methods

**Stimuli.** An algorithm that attempts to classify whether or not two sentences share the same thematic role assignments, based on any subset of hidden units in a LLM's, can solve the task based on simple "tricks". For example, if some units represent the position of each noun in the sentence, and some units represent the existence of the word "by", then two sentences have the same meaning if the nouns are in opposite orders across the two sentences, and only one has the word "by". Therefore, for Experiment 2, we created a more challenging stimulus set, consisting of ditransitive sentences with a variety of structures (e.g., "the man gave the milk to the woman", "it was the woman that was given the milk by the man"), where no global "trick" can be applied to infer whether sentences shared meaning or not. A single stimulus set used 12 structures, each with two versions having opposite agent-patient assignments, for a total of 24 sentences. Structures varied in whether they were active or passive, used a double- or prepositional-object construction (DO vs. PO), and were in simple or cleft form (with a subject, direct object, or indirect object focus). We generated 50 sets of such 24 sentences, for a total of 1,200 sentences. Example stimuli are provided in **Table 3**.

**Table 3.** Example stimuli for Experiment 2.[a]

| Active / Passive | DO / PO | Simple / cleft | Sentence |
|---|---|---|---|
| Active | DO | Simple | **The man** gave the woman the milk. |
| | | Cleft (subject) | It was **the man** who gave the woman the milk. |
| | | Cleft (dir. object) | It was the milk that **the man** gave the woman. |
| | PO | Simple | **The man** gave the milk to the woman. |
| | | Cleft (subject) | It was **the man** who gave the milk to the woman. |
| | | Cleft (dir. object) | It was the milk that **the man** gave to the woman. |
| | | Cleft (indir. object) | It was the woman who **the man** gave the milk to. |
| Passive | DO | Simple | The woman was given the milk by **the man**. |
| | | Cleft (indir. object) | It was the woman who was given the milk by **the man**. |
| | PO | Simple | The milk was given to the woman by **the man**. |
| | | Cleft (dir. object) | It was the milk that was given to the woman by **the man**. |
| | | Cleft (indir. object) | It was the woman who the milk was given to by **the man**. |

[a] The agent in all sentences is shown in bold (the man). Each stimulus set consisted 12 more sentences (one more version of each structure) with the opposite thematic role assignment (in this example, woman as agent). Abbreviations: dir = direct; indir = indirect.

**LLMs: hidden units.** Prior to training a classifier algorithm on pairs of sentences, we performed the same analysis as in Experiment 1, computing cosine similarities for every pair of sentences within each stimulus set based on the pattern of activity distributed across all hidden units in each layer. We tested whether similarities for pairs that shared thematic role assignments were higher than for pairs with opposite assignments, splitting the analysis by whether the respective structures of the two sentences in a pair differed in 0, 1, 2, or 3 of the syntactic features described above. An example pair of sentences with 0 changes and the same thematic role assignments: "it was the man who gave the woman the milk" and "it was the milk that the man gave the woman"; both are active, use the direct object construction, and have a cleft, but each cleft has a different focus. A pair of sentences that differ in 1 feature are "the man gave the woman the milk" and "it was the man who gave the woman the milk", as they are both active and DO, but the second sentence has a cleft. A pair of sentences that differ in 2 features are "the man gave the woman the milk" and "the milk was given to the woman by the man", as the second sentence is passive and has a cleft. A pair of sentences that differ in 3 features are "the man gave the woman the milk" and "it was the milk that was given to the woman by the man", as the second sentence is passive, has a cleft, and a prepositional object construction. There are other ways to split our sentence pairs based on their syntactic differences, and we do not claim that our distinctions are more cognitively relevant; we chose this scheme because it reflects how we constructed the stimuli. We emphasize that this analysis is exploratory in nature. *p*-values were Bonferroni-corrected for multiple comparisons across layers, separately for each LLM.

For our main analysis, for each layer in each LLM, we trained a support vector machine (SVM) algorithm to distinguish between "same meaning" vs. "different meaning" sentence pairs (i.e., pairs with matching vs. opposite thematic role assignments). We use a linear classifier, consistent with previous research that has found conceptual information in hidden layers with this approach (Lee et al., 2024; Nanda et al., 2023; Ramezani et al., 2025; Seyffarth et al., 2021) (also see Discussion). To this end, for each pair of sentences, their respective distributed representations were subtracted and the resulting difference was submitted as a training instance to the algorithm (we also attempted to train the algorithm on the concatenated representations of two sentences in a pair; results were similar or weaker). In a 66-fold cross-validation, a separate SVM was trained on each possible subset of 10 of the 12 sentence structures and tested on the remaining two held-out structures.

Specifically, for each fold, training data were generated from 1,000 sentences: 20 out of the 24 sentences in each stimulus set (10 structures × 2 possible agent-patient assignments). Within each of our 50 stimulus sets, we used nearly all possible sentence pairs, excluding pairs consisting of two versions of the exact same structure (e.g., passive + prepositional object + cleft with a direct object focus), because those pairs always had opposite thematic role assignments. Overall, each fold included 9,000 training pairs. Because the held-out test data included 2 structures, each with two versions (differing in thematic role assignments), there were 4 test pairs per stimulus set, for a total of 200 pairs per fold (again, excluding pairs consisting of two versions of the same structure).

For each layer of each LLM, binary classification accuracy was tested against chance performance (0.5) using a one-sample *t*-test. *p*-values for Bonferroni-corrected for multiple comparisons across layers, separately within each LLM.

**LLMs: attention heads.** For each attention head in each layer, we studied attention patterns assigned between content words in each sentence. Namely, we extracted each sentence's attention weights between every pair of the following words: subject (which, depending on sentence structure, was the agent or patient), indirect object (patient or agent), verb, and indirect object; we excluded attention from the verb to the direct object and vice versa because these involved neither agent nor patient. Using these vectors of 10 attention weights per sentence, the same SVM analyses described above were conducted to classify sentence pairs with matching vs. opposite thematic role assignments. However, here we used a concatenation of the two vectors of a pair of sentences rather than their difference, because this representation resulted in better performance and we wanted to give LLMs the best opportunity to succeed. Because the relative position of words varied across sentence structures, this analysis sometimes necessitated "forward-looking" attention (i.e., from a previous word to a future word). However, attention in most of our models is only backward-looking (unidirectional), so our analysis was limited to BERT, which has bidirectional attention. *p*-values were Bonferroni-corrected for multiple comparisons across all attention heads and layers (12 heads per layer × 12 layers = 144 tests).

**Human judgments.** We recruited two samples (121 and 120 participants, respectively) through UCLA's participant recruitment system for an online experiment. Participants rated pairs of sentences for their similarity as in Experiment 1. In the first sample, each participant judged only two critical pairs of sentences: in one pair, the two sentences shared the same thematic role assignments, and in the other they had opposite assignments. These two trials were from different stimulus sets. Due to a coding error, in this first sample, we only sampled 12 out of the 50 stimulus sets. Furthermore, due to another error, each participant rated one pair in which the two sentences differed in 3 syntactic features (passive-DO-simple vs. active-PO-cleft), and another pair in which they differed in 2 syntactic features (active-simple vs. passive-cleft, both with a PO). Five filler trials were included as in Experiment 1.

The second sample of participants rated four critical pairs of sentences alongside 18 filler trials, and the full stimulus set was sampled across participants. In two pairs—one with shared and one with opposite assignments—the two sentences differed in a single syntactic feature (for some participants, passive-simple with DO vs. PO; for other participants, active with simple vs. cleft form, either both with a DO or both with a PO). In the two remaining pairs—again, one with shared and one with opposite assignments—the two sentences differed in two syntactic features (for some participants, passive-simple vs. active-cleft, both with a PO; for other participants, active-DO vs. passive-PO, both in simple form).

We *z*-scored similarities within each participant. We then tested whether similarities for sentence pairs that shared thematic role assignments were higher than for pairs with opposite assignments, splitting the analysis by whether the respective structures of the two sentences in a pair differed in 1, 2, or 3 of the syntactic features. For pairs differing in one feature, we fitted a linear, mixed-effects model predicting sentence similarity from condition (shared vs. opposite thematic roles), with random intercepts by participant and by stimulus set: Similarity ~ Condition + (1 | Participant) + (1 | Set). For pairs differing in two features, we fitted a similar model but without a random intercept by participant, because some participants (namely, those from the first sample) only saw a single sentence pair of this format. For pairs differing in three syntactic features, we did not include a random intercept by participant for the same reason; the model including a random intercept by stimulus set did not converge, so it was reduced to a fixed effects model predicting similarity rating from condition. *p*-values were Bonferroni-corrected across these 3 models.

## Results

**LLMs: Sentence similarity**. When considering sentence representations distributed across all hidden units within a given layer, we did not find evidence of robust representation of thematic roles (**Figure 4**). Only a minority of cases showed the expected pattern, whereby pairs of sentences that share thematic role assignments are more similar than pairs with opposite assignments. For all LLMs, we also observed a strong influence of syntax on sentence similarity: sentence pairs differing in one syntactic feature were overall more similar than pairs differing in two syntactic features, which were overall more similar than pairs differing in three syntactic features; this pattern held regardless of whether pairs shared the same thematic role assignments or had opposite assignments.

In BERT, for sentence pairs differing in one, two, or three syntactic features, similarity did not vary as a function of whether the pair had the same thematic role assignments (one feature: $z$=1.94, $p$ =.21, Bonferroni-corrected here and below; two features: $z$ = .52, $p$ =1; three features: $z$ = .0077, $p$ = 1; **Figure 4A**). Pairs with no differences in syntactic features showed a pattern that is opposite of what would be expected from a representation that is sensitive to thematic roles: pairs with opposite thematic role assignments were more similar than pairs which shared thematic role assignments ($z$=17.05, $p$<$10^{-63}$).

In GPT2, for sentence pairs differing in one syntactic feature, pairs which shared thematic role assignments were more similar than pairs with opposite assignments, consistent with sensitivity to thematic roles ($z$=5.06, $p$<10$^{-5}$; **Figure 4B**). However, pairs with no differences in syntactic features, as well as pairs differing in 2 or 3 features, showed the opposite pattern (no difference: $z$=11.05, $p$<10$^{-27}$; two features: $z$=2.78, $p$=.02; three features: $z$=4.46, $p$<10$^{-4}$).

In Llama2, for sentence pairs differing in one syntactic feature, pairs which shared thematic role assignments were more similar than pairs with opposite assignments, consistent with sensitivity to thematic roles ($z$=3.52, $p$=.001; **Figure 4C**). However, pairs with no differences in syntactic features showed the opposite pattern ($z$=6.69, $p$<10$^{-10}$). For pairs that differed by 2 or 3 features, similarity did not vary as a function of whether the pair had the same meaning (two features: $z$=.71, $p$=1; three features: $z$=.15, $p$=1).

Finally, the results for Persimmon mirrored those of Llama2. For sentence pairs differing in one syntactic feature, pairs which shared thematic role assignments were more similar than pairs with opposite assignments, consistent with sensitivity to thematic roles ($z$=4.70, $p$<10$^{-4}$; **Figure 4D**). However, pairs with no differences in syntactic features showed the opposite pattern ($z$=6.58, $p$=10$^{-10}$). For pairs that differed by 2 or 3 features, similarity did not vary as a function of whether the pair had the same meaning (two features: $z$=.20, $p$=1; three features: $z$=.52, $p$=1).

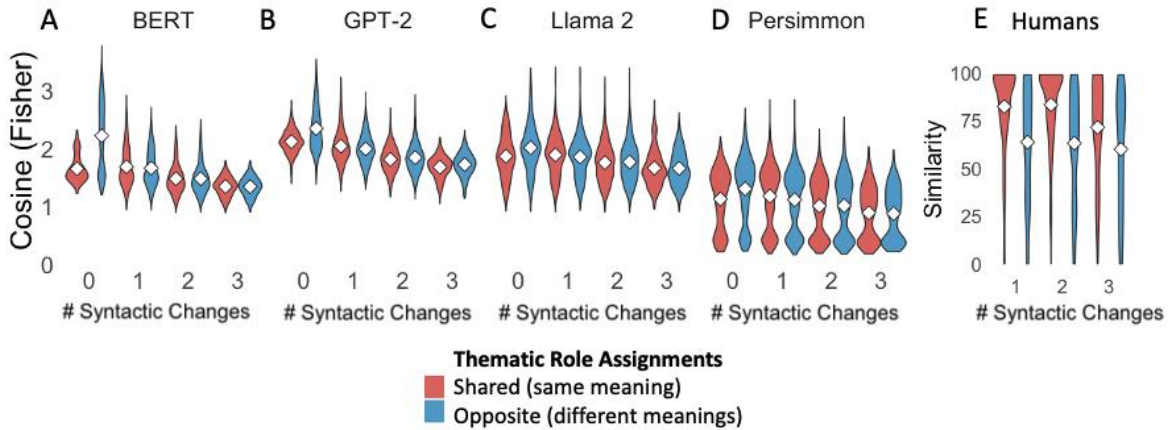

**Figure 4:** Results of Experiment 2, showing similarity patterns between sentence pairs that have shared (red) or opposite (blue) thematic role assignments, for (A) BERT, (B) GPT2, (C) Llama2, (D) Persimmon, and (E) humans. For LLMs, data are shown for the last hidden layer, and cosine similarities are Fisher-transformed. Violin plots show the distribution over sentence pairs. White diamonds show the mean.

**Human Judgments.** Human similarity judgments showed higher sensitivity to thematic role information compared to LLMs (**Figure 4E**): they rated sentence pairs which shared thematic role assignments as more similar than pairs with opposite assignments, regardless of whether the two sentences in a pair differed in a single syntactic feature ($z$=6.25, $p$<$10^{-9}$), in two features ($z$=7.90, $p$<$10^{-14}$), or in three features ($t_{(118)}$=3.04, $p$=0.004; this test was between- rather than within-participants). For pairs differing in either one or two syntactic features, 70.3% of participants provided ratings in this direction (CI$_{95\%}$ = [62%, 78.5%], based on 5,000 bootstrap samples) (such a percentage could not be calculated for pairs differing in 3 syntactic features, because no participant saw more than one sentence pair of this format; see Methods).

**SVM: hidden units**. Classification accuracies of sentence pairs with shared vs. opposite thematic role assignments (i.e., "same" vs. "different" meanings), for each layer in each LLM, are shown in **Figure 5**. Many SVM results significantly exceeded chance level (0.5), but with a relatively small effect size, mostly below 0.6 accuracy. The highest classification accuracy was obtained for GPT2, layer 5. For this layer, accuracy exhibited high variance, ranging from 0.995 to 0.32 across folds, with the latter suggesting overfitting; and nearly 35% of the folds showed chance accuracy or lower (layer 1 of Llama2 also exceeded 0.6 accuracy, and similarly exhibited high variance in accuracy across folds).

SVM performance in this GPT2 layer often (but not always) lagged behind that of humans. As a reminder, human similarity ratings were significantly higher for sentence pairs with matching vs. opposite thematic roles; for sentence pairs differing in either one or two syntactic features, 70.3% of participants provided similarity ratings in this direction, i.e., human implicit "classification accuracy" was 0.703. For comparison, in layer 5 of GPT2, the 95% confidence interval of classification accuracy was [0.594, 0.667] (5,000 bootstrap samples across folds). SVM performance, broken down by the number of syntactic features that differed between sentences in a pair, was as follows: 0 features, 0.638; one feature, 0.658 (on the particular folds that tested the same structures that humans rated: 0.5, 0.635, and 0.985); two features, 0.624 (on the particular folds that tested the same structures that humans rated: 0.40, 0.615, 0.62, and 0.77); and three features, 0.542 (on the particular fold that tested the same structures that humans rated: 0.32). Of all folds that tested sentence pairs differing in one syntactic feature, 35.7% had accuracy significantly lower than humans, and 21.4% had accuracy significantly higher than humans (two-tailed, two-proportion $z$-test, Bonferroni corrected for multiple comparisons). Of the folds that tested sentence pairs differing in two syntactic features, 28% had accuracy significantly lower than humans, and 4% had accuracy significantly higher than humans.

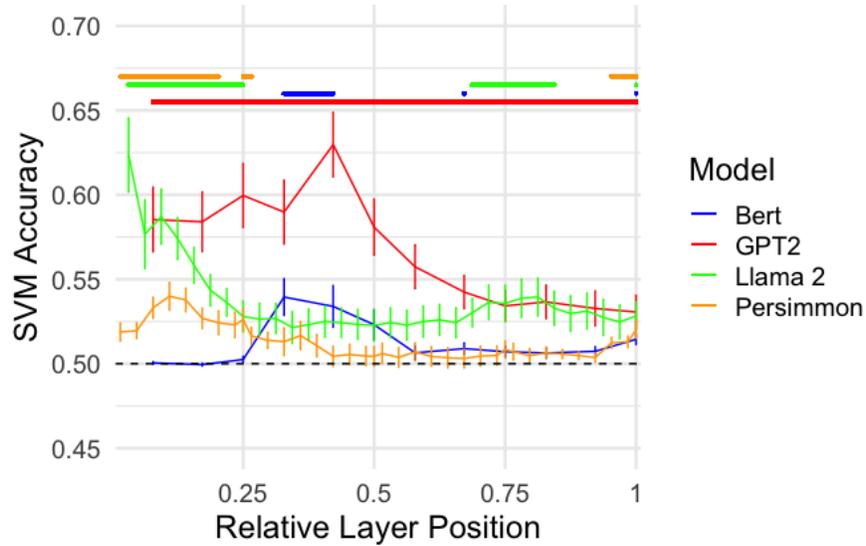

**Figure 5**: Classification accuracies for SVMs, trained on distributed activations across hidden units to predict whether two sentences had shared vs. opposite thematic role assignments (i.e., "same" vs. "different" meanings). A separate SVM was trained per layer, and the *x*-axis shows relative layer position. Different curves correspond to different LLMs. Error bars show standard errors of the mean. A black, dotted line denotes chance classification. Colored, horizontal lines represent where classification accuracy is significantly above chance.

**SVM: attention heads.** Classification accuracies of sentence pairs with shared vs. opposite thematic role assignments (i.e., "same" vs. "different" meanings), for each attention head in each layer of BERT, are shown in **Figure 6**. Several heads had high accuracy, and we highlight head 5 in layer 11, which had the highest accuracy: 0.79. SVM performance, broken down by the number of syntactic features that differed between sentences in a pair, was as follows: 0 features, 0.801; one feature, 0.794 (on the particular folds that tested the same structures that humans rated: 0.755, 0.79, and 0.79); two features, 0.792 (on the particular folds that tested the same structures that humans rated: 0.77, 0.795, 0.795, and 0.84); and three features, 0.784 (on the particular fold that tested the same structures that humans rated: 0.82). Whereas only two folds had accuracy that significantly exceed that of humans, the 95% confidence interval of classification accuracy across folds was [0.784, 0.801] (based on 5,000 bootstrap samples), compared to our proxy of human implicit "classification accuracy" of 0.703.

To characterize this head's function (**Figure 7**), we contrasted its attention patterns (1) from the verb to the agent vs. patient; (2) from the direct object to the agent vs. patient; and (3) from agent to the patient vs. vice versa. Each comparison was carried out in a linear, mixed-effects model. Attention weights, which are restricted between [0,1], were logit-transformed (this did not change the results) and modeled with a fixed effect of direction (towards the agent vs. patient) and random intercepts and slopes by stimulus set and by structure: Attention ~ Direction + (1 + Direction | Set) + (1 + Direction | Structure)  (based on the 12 structures in Table 3). The verb ($t_{(51.48)}$=13.76, $p$<10$^{-16}$) and direct object ($t_{(36.48)}$=5.18, $p$<10$^{-5}$) both allocated more attention to the agent than to the patient, and patients directed more attention to agents than vice versa ($t_{(35.60)}$=7.76, p<10$^{-8}$). These patterns held across most sentence structures regardless of the grammatical positions of agents and patients. Therefore, they robustly reflect thematic roles, independent of syntax.

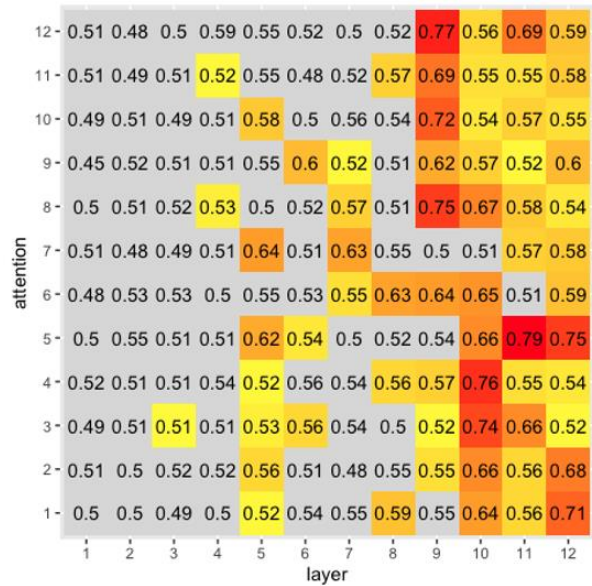

**Figure 6:** Classification accuracies for SVMs, trained on patterns of attention in BERT to predict whether two sentences had shared vs. opposite thematic role assignments (i.e., "same" vs. "different" meanings). A separate SVM was trained per attention head, and each cell in the matrix shows the result for one head (row) in one layer (column). Accuracies significantly above chance are colored (Bonferroni-corrected for multiple comparisons across all heads).

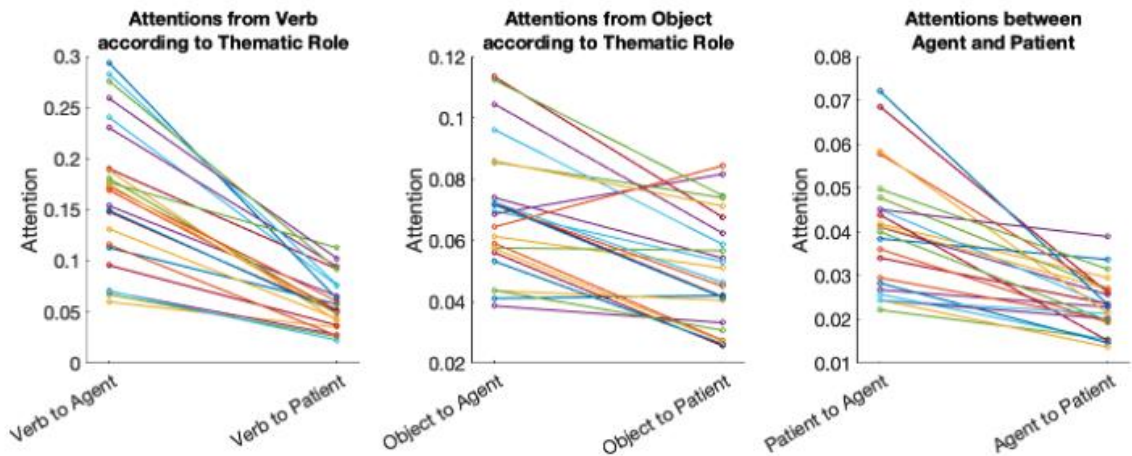

**Figure 7:** Attention patterns between different thematic roles, for BERT's head 5 in layer 11. Each line corresponds to one for 24 sentence types (12 structures × thematic role assignments), averaged across stimulus sets.

## Discussion

This study asked whether Large Language Models (LLMs) understand sentences in the minimal sense of representing "who did what to whom". In Experiment 1, we found that the overall geometry of LLM distributed activity patterns failed to capture this information: similarities between sentences reflected whether they shared syntax more than whether they shared thematic role assignments. Human judgments, in contrast, were strongly driven by this aspect of meaning. In Experiment 2, we found limited evidence that thematic role information was available even in a subset of hidden units. Whereas activity patterns in subsets of hidden units often allowed for significant classification of whether sentence pairs had shared vs. opposite thematic role assignments, the effect sizes were small; even the best-performing case appeared to lag behind humans, and its representation of thematic roles did not seem robust across syntactic structures. However, thematic role information was reliably available in a large number of attention heads, demonstrating LLMs have the capacity to extract thematic role information. In some cases, information present in attention heads descriptively exceeded human performance.

These results provide an important characterization of the ability of LLMs to understand language: thematic roles are tightly linked to understanding linguistic input as such, in contrast to aspects of comprehension such as common-sense reasoning or logical inference which, while frequently studied in LLMs, reflect non-linguistic aspects of thinking (Mahowald, Ivanova, et al., 2024). Even in our relatively simple semantic task of mapping sentence structure onto thematic roles, LLMs do not give meaning the prominent role that humans do—at least not in their hidden units—despite possessing their ability to extract this information. If we take the distributed patterns of activity across hidden units to be a representation of a sentence, as is often done in studies comparing LLMs to human data (Antonello et al., 2021; Caucheteux and King, 2022; Schrimpf et al., 2021; Tang et al., 2024; Tuckute et al., 2024), then our findings suggest that training LLMs on word prediction is sufficient for learning what thematic roles are, but perhaps not for representing them in a human-like way.

Our findings are consistent with the broader claim that the success of LLMs in syntactic processing does not guarantee similar success in semantic processing (Weissweiler et al., 2022). Indeed, despite the impressive syntactic capabilities of LLMs (e.g., Manning et al., 2020; McCoy et al., 2023; Wilcox et al., 2021, for reviews, see: Chang and Bergen, 2024; Futrell and Mahowald, 2025; Linzen and Baroni, 2021) prior work has demonstrated that LLMs trained on word prediction alone have limited understanding (though this has often been evaluated on non-linguistic aspects of understanding): LLMs struggle with tracking the state of entities in a text (Kim and Schuster, 2023), sometimes refer to entities that do not exist (Schuster and Linzen, 2022), and make predictions that are only weakly sensitive to event roles (Ettinger, 2020).

As stated in the introduction, LLMs can perform thematic role assignment if they are directly fine-tuned on this task. However, our study asked whether robust representations of thematic roles could result from training on word prediction exclusively, in order to understand how a general linguistic objective affects LLM semantic capabilities. Our work thus adds to the existing endeavor of characterizing models that lack fine-tuning (for a review, see: Chang and Bergen, 2024). Probing such models is crucial as they are the ones that are commonly compared to human behavior and brain activity. For instance, one important interpretation of our findings is that whatever brain activity can be predicted from the hidden activations of LLMs, it likely does reflect thematic roles.

At the same time, this study adds to a growing literature suggesting that attention heads—a critical component of LLMs that is distinct from hidden units—contain important linguistic information. Previous work has found that attention heads and circuits of attention can identify parts of speech and syntactic dependencies (Clark et al., 2019; Manning et al., 2020; Wang et al., 2022) and, moreover, that those computations are influenced by semantic information (McGee and Blank, 2024). Here, we demonstrate that thematic role information is also reliably captured by attention heads. The relationship between heads that identify syntactic relations and those that identify semantic relations remains an open question for future studies. Other work, focusing on BERT's attention heads, has shown that they tend to be "overparametrized", with many heads attending to the same information (e.g., noun-pronoun or verb-subject links), and that disabling some heads can even improve performance (Kovaleva et al., 2019). Our findings are consistent with these claims, as we found many attention heads that appear to be sensitive to thematic roles. Whereas our attention head analysis focused on BERT, because our evaluation method required a model with bidirectional attention, future work may test whether heads in unidirectional models also extract thematic role information. Future work may also consider the casual influence of thematic role information in attention heads (or circuits of attention; Wang et al., 2022) on next word prediction and LLM behavior more broadly.

Our results from human participants show that they can rely on thematic roles to determine the similarity of sentences, even when the only cues to thematic roles are grammatical positions (rather than, e.g., plausibility), and without explicit instructions to consider sentence structure. This was especially apparent in Experiment 1, where our sentence stimuli were rather simple to understand. Results from Experiment 2 were somewhat weaker, as some of the sentences had more complex structures that are harder to parse, and some sentence pairs differed in three syntactic features. In those cases, the classification accuracy for attention heads exceeded human performance. Therefore, whereas our analyses of hidden units reveal one difference between humans and LLMs, wherein LLMs represent thematic role information more weakly, our analysis of attention heads demonstrates a difference in the opposite direction.

Our study relies on probing the internal representations of LLMs as a means to reveal what they know. Probing internal representations is a different approach from studying the behavior of LLMs by prompting them with questions and characterizing their output. One may simply feed LLMs a sentence and ask "who is the agent?", or feed two sentences and ask whether they have matching vs. opposite thematic roles. However, such prompting makes several undesirable assumptions about LLMs: that they understand the meaning of words like "agent", that they understand how the question being asked of them relates to the sentences asked about, etc. Studying internal representations circumvents these additional "task demands" (Hu et al., 2024; Hu and Frank, 2024; Hu and Levy, 2023). In the Introduction, we further justify our use of probing instead of prompting, and we believe that our study of internal representations provides a more informative and detailed characterization of LLMs than any results that would be obtained with prompting.

We recognize that our paradigms for testing the LLMs and the human participants differ in important ways. In contrast to LLMs, we did not probe the internal representations of our human participants, as this would require neuroimaging that is beyond the scope of this experiment; instead, we "prompted" the participants with explicit questions about sentence similarity. Nonetheless, previous work has shown that thematic role information is indeed present in human neural representations, and meaning can be classified based on patterns of brain activity using linear classifiers similar to the ones we used here (Frankland and Greene, 2015; Wang et al., 2016).

Whereas our analyses of thematic role information in LLMs is extensive, it could not be comprehensive. For instance, we used a specific similarity metric (cosine) and rather simple classifiers (linear SVMs). It is possible that thematic role information is present non-linearly within the hidden representations of LLMs (see Hernandez et al., 2023). Previous research, however, has supported the "linear subspace hypothesis": that much conceptual information is linearly represented in hidden representations of LLMs (Bolukbasi et al., 2016; Vargas and Cotterell, 2020). Indeed, previous work has found that several complex concepts are present linearly in the embedding space, including gender bias (Bolukbasi et al., 2016; Vargas and Cotterell, 2020), truthfulness (Li et al., 2024; Marks and Tegmark, 2023), space and time (Gurnee and Tegmark, 2023), sentiment (Tigges et al., 2023), plausibility (Lee et al., 2024), and world states (Nanda et al., 2023). Therefore, if thematic role information is available in hidden layers, it would have to differ in its representational format from these other semantic features. Additionally, we evaluated sentence representations from one token (extracting activations at [CLS] for BERT, or '.' for the other LLMs; but see Supplementary Materials). It is possible that thematic role information may be present in other tokens, and we leave such investigations for future work. However, we emphasize that most LLMs use unidirectional attention, so the representation of any token other than the last one cannot capture the entire sentence; this leads to problems when studying structurally diverse materials such as those we used in Experiment 2, where different thematic roles can appear very early or very late in the sentence. Moreover, previous work has found that conceptual information is present in sentence representation tokens like [CLS] (e.g., Ramezani et al., 2025; Seyffarth et al., 2021).

Critically, we do not suggest that LLMs are principally incapable of representing thematic roles in more human-like ways. Even though we studied LLMs of varying sizes, and found that increasing model size does not appear to improve the representation of thematic roles across hidden units, it is possible that other architectures, or other training corpora, could lead models to generate such representations. In addition, models exposed to non-linguistic training, e.g., reinforcement learning from human feedback, may process thematic roles differently than those trained only on word prediction.

Nonetheless, our findings emphasize that it is vital to test the ability of LLMs to process language "per se" (cf. exhibit non-linguistic thinking), and to do so using carefully designed and controlled materials inspired by psycholinguistics. Such studies are required for ensuring that the seemingly meaningful text that LLMs generate reflects comprehension rather than non-linguistic

"tricks" that have little to do with human language processing (McCoy et al., 2019). In the case of thematic roles, this approach is important because, in natural texts, these roles might be assigned based on heuristics such as plausibility (Mahowald et al., 2023). Whereas humans can use such information, in some cases they can also process thematic roles based on sentence structure alone. Our findings demonstrate that LLMs can extract such information via their attention heads, but when it comes to their hidden units—commonly studied as the "ultimate" internal representation of LLMs that are compared to human cognitive representations—thematic role information exerts a relatively weak influence.

## Acknowledgements

We would like to thank the audiences of the 37[th] Annual Conference on Human Sentence Processing  and the 46[th] Annual Meeting of the Cognitive Science Society for helpful comments on this work. We also thank our friend, colleague, and co-author, Dr. Bryor Snefjella, who passed away in March 2023, and is dearly missed.